\documentclass[12pt, letter]{article}
\usepackage{geometry,epsfig,multirow,graphicx,subfigure,natbib,url}
\usepackage{amsmath,amssymb,theorem,float,crop,algorithmic,algorithm,bbm,bm,mathrsfs}

\newcommand{\VAR}{{MCEM-VAR }}
\newcommand{\ARS}{{MCEM-ARS }}
\newcommand{\ARSID}{{MCEM-ARSID }}
\newcommand{\FEAT}{{FEAT-ONLY }}

\newcommand{\commentout}[1]{}
\newcommand{\parahead}[1]{\vspace{0.07in}\noindent {\bf #1:}}
\newcommand{\norm}[1]{\lVert#1\rVert}

\newcommand{\myItemizeBegin}{
 \begin{list}{$\bullet$}
  { \setlength{\itemsep}{1pt}
     \setlength{\parsep}{0pt}
    \setlength{\topsep}{0pt}
    \setlength{\partopsep}{0pt}
    \setlength{\leftmargin}{2em}
    \setlength{\labelwidth}{1.5em}
    \setlength{\labelsep}{0.5em} } }
\newcommand{\myItemizeEnd}{\end{list}}

\begin{document}

\title{Parallel Matrix Factorization for Binary Response}
\author{
Rajiv Khanna, Liang Zhang, Deepak Agarwal, Beechung Chen \\
Yahoo! Labs \\
4401 Great America Pkwy, Santa Clara, CA\\
\{krajiv,liangzha,dagarwal,beechun\}@yahoo-inc.com}


\maketitle

\begin{abstract}

Predicting user affinity to items is an important problem in 
applications like content optimization, computational
advertising, and many more. While bilinear random effect models (matrix factorization) provide
state-of-the-art performance when minimizing RMSE through
a Gaussian response model on explicit ratings data, applying it to
imbalanced binary response data presents additional challenges that we
carefully study in this paper. Data in many applications usually consist of users'
implicit response that are often binary -- clicking an item or not; the
goal is to predict click rates (i.e., probabilities), which is often
combined with other measures to calculate utilities to rank items at
runtime of the recommender systems.  Because of the implicit nature, such data are usually much
larger than explicit rating data and often have an imbalanced distribution
with a small fraction of click events, making accurate click rate
prediction difficult. In this paper, we address two problems. First, we
show previous techniques to estimate bilinear random effect (BIRE) models with binary data
are less accurate compared to our new approach based on 
adaptive rejection sampling, especially for imbalanced response. Second, we develop a parallel bilinear random effect model fitting framework using Map-Reduce paradigm that scales to
massive datasets. Our parallel algorithm is based on a ``divide and
conquer'' strategy coupled with an ensemble approach. Through
experiments on the benchmark MovieLens 1M data, a small Yahoo! Front Page Today Module data set, and a large Yahoo! Front Page Today Module data set that contains 8M users and 1B binary observations, we show that careful
handling of binary response as well as identifiability issues are needed to achieve good performance for
click rate prediction, and that the proposed adaptive rejection
sampler and the partitioning as well as ensemble techniques significantly improve model performance.


\end{abstract}




\section{Introduction} \label{sec:intro}

Personalized item recommendation is an important task in many web
applications, such as content optimization~\citep{agarwal08nips},
computational advertising~\citep{Broder08}, and others.  Such
systems recommend a set of items like article links, ads,
product links, etc for each user visit; users respond by clicking and/or
engaging in other activities post-click.  
Personalizing such recommendations based on the user's demographic information and browsing history, typically leads to better user engagement and profit for organizations. 
Accurate
prediction of the probability of a user clicking an item, is an important input to facilitate such personalization.

For a user $i$ and an item $j$, let $p_{ij}$ be the probability of user $i$ clicking item $j$.
If we let $r_j$ denote the
utility of clicking item $j$ (e.g., the ad revenue associated with a click
on item $j$), the system may rank items based on some function of
$r_j$ and $p_{ij}$ to maximize the utility. In computational advertising for
instance, ranking is based on the expected revenue $r_{j}p_{ij}$. Click
probabilities are usually estimated through a statistical model
trained on past click data. Intuitively, we can think of the click
data as a binary matrix $\bm{Y}$, such that entry $y_{ij} = 1$ if user $i$
clicked item $j$, and $y_{ij} = 0$ if user $i$ viewed but did
not click on $j$.  We note that $\bm{Y}$ is a highly incomplete matrix with
many unobserved entries since each user usually views a small number
of items. The goal is to predict $p_{ij}$ for unobserved $(i,j)$
pairs. We also note that in many web applications the click-rates are small giving
rise to highly imbalanced binary response data.  

\subsection{Background and Literature}
The problem of personalized item recommendation described above is closely related to a rich literature on recommender systems and collaborative filtering. An overview can be seen in \cite{adomavicius2005toward}. Recommender systems are algorithms that model user-item interactions to facilitate the process of personalized item recommendations. There are two types of approaches that are widely used in recommender systems: \em content-based\em approaches and\em collaborative filtering\em. The content-based approaches use only user and item covariates to model the user-item interaction. Collaborative filtering approaches model user-item interactions by user's past response alone, no covariates are used. However, in real recommender systems we often observe both ``warm-start" and ``cold-start" scenarios: ``Warm-start" means we have past observations from this user/item so that both the past responses and covariates can be used in modeling. ``Cold-start" scenario represents when a new user/item comes to the system; hence we do not have any past responses but may still have the covariates. To handle both scenarios a hybrid approach that combines content-based and collaborative filtering is often used in recommender systems.

Nearest-neighbor methods are widely used in collaborative filtering (e.g. \cite{sarwar2001item}, \cite{wang2006unifying}). They are very popular in large-scale commercial systems, such as \cite{nag2008vibes}, \cite{linden2003amazon}. The basic idea of the nearest-neighbor methods is to compute item-item similarity or user-user similarity from Pearson correlation, cosine similarity or Jaccard similarity of the responses of a pair of users/items. Then for each unobserved user-item pair, the prediction is simply a weighted average of the set of nearest neighbor's responses, and the weights come from the similarity measures. More recently, \cite{agarwal2011modeling} proposed a Bayesian hierarchical modeling approach to model the item-item similarity in a more principled way.

Since the Netflix challenge \citep{nworkshop}, the SVD-style matrix factorization methods have been well known to provide state-of-the-art performance for recommender problems. In this paper, we shall refer to the class of matrix factorization models as bilinear random effects (\textbf{BIRE}) models. A
theoretical perspective of this problem was first provided in
\cite{srebro2005}. \cite{koren07,nworkshop} have successfully used this strategy in collaborative filtering applications. \cite{ruslan:icml08b,ruslan08a} formulate a probabilistic framework using a hierarchical random-effects model where the user and item factors are multivariate
random-effects (factor vectors) that were regularized through zero-mean multivariate
Gaussian priors. These papers used bilinear random effects as a tool to solve pure collaborative filtering problems; they do not
work well in applications with significant cold-starts which is commonplace in several applications. 

In light of the ``cold-start" problem, a desirable approach would be to have a model that provides predictive accuracy as
BIRE for warm-start scenarios but fallbacks on a
feature-based regression model in cold-start scenarios.
Incorporating both warm-start and cold-start scenarios simultaneously in collaborative filtering 
is a well studied problem with a rich literature.  Several methods that
combine content and collaborative filtering have been studied. For instance,
\cite{fab97} present a recommender system that computes user similarities based
on content-based profiles. In \cite{claypool99}, collaborative
filtering and content-based filtering are combined linearly with
weights adjusted based on absolute errors of the
models. In \cite{melville02}, content based models are used to fill up
the rating matrix followed by recommendation based on similarity (memory) based methods \citep{BreeseHK98}. 
In \cite{good99,park06}, \textit{filterbots} are used to
improve cold-start recommendations. \cite{schein03} extend the
aspect model to combine the item content with user ratings under a single
probabilistic framework.  In \cite{agarwal2009regression,matchbox}, a principled solution is proposed by using linear model regression priors on the user and item random effects through features. This is generalized in \cite{gmf} so that the regression priors can be non-linear functions which further improves the model predictive accuracy. These solutions based on incorporating features into random effects itself are
superior than the classical methods of dealing with warm-start and cold-start
scenarios simultaneously. 

\subsection{Statistical Challenges}
The basic idea of the bilinear random effect (BIRE) models is to approximate the response $y_{ij}$ from user $i$ and item $j$ by an inner product of the user random effect $\bm{u}_i$ and the item random effect $\bm{v}_j$. For Gaussian responses, the loss function is usually RMSE, i.e. $\min \sum_{ij} (y_{ij} - \bm{u}_i^\prime \bm{v}_j)^2$ over observed $(i,j)$ pairs. Due to a preponderance of missing entries in $\bm{Y}$, the user/item random effects have to be regularized further to avoid over-fitting the training data. This is usually done by constraining latent profiles through the $L_2$ norms or equivalently assuming
zero-mean Gaussian priors on the user/item random effects. To handle cold-start scenarios, instead of zero-mean Gaussian priors, covariates-based regression priors can be used on the user/item random effects, and both random effects and the regression parameters are estimated simultaneously through a Monte Carlo EM (MCEM) algorithm \citep{booth99mcem}. \cite{agarwal2009regression,gmf} show that this modeling framework gives state-of-the-art performance, especially for Gaussian user-item interaction responses. However, in some real recommender systems we often observe the following two major challenges:
\begin{itemize} 
\item {\bf Imbalanced binary response:} Many web applications depend on
  implicit user feedback that are in the form of events like clicks.
  Further, these events are usually rare; this gives rise to imbalanced binary response data.
  Accurate estimation of probabilities with imbalanced binary response is known to be a
  difficult problem~\citep{owen} even in the case of ordinary logistic regression; performing such
  estimation for elaborate BIRE models introduces additional challenges that have not been carefully studied before.

\item {\bf Large scale data:} As every
  display of an item to a user generates an observation, data
  collected from these applications is massive for large systems such as major websites.
  In fact, the entire data often does not fit into memory and 
  resides in large distributed data clusters. Scalable model fitting using distributed
  computing paradigm like Map-Reduce \citep{dean2008mapreduce} is an attractive option. However, the EM algorithms
  used to fit such BIRE models require sequential processing of data and are not
  directly amenable to computing in a Map-Reduce paradigm. 
  Therefore, scalable model fitting for such massive datasets is still a challenge.
\end{itemize} 
We address both of the challenges mentioned above in this paper. First,
although a variational approximation method has been proposed
in~\cite{agarwal2009regression} to perform approximate sampling of the user/item
random effects for binary response data in the E-step of the Monte Carlo EM
model fitting procedure, we show such an approximation does not give optimal
prediction accuracy, especially for imbalanced binary response. By
using an adaptive rejection sampling (ARS) technique to sample the
factors exactly in the E-step, we can significantly improve prediction accuracy. We
also find model identifiability issues when dealing with imbalanced
binary response that hurts model accuracy; we show that this can be
handled effectively in our framework by enforcing a few additional
constraints on the random effects. To our surprise, we find the
variational approximation deteriorates as the response gets rare and
it happens because the random effect estimates shrink more compared to the 
exact ARS sampler.   

To handle model fitting for massive datasets, we develop a parallel BIRE model fitting framework
based on Map-Reduce that fits accurate models in a
scalable way. Our method is based on two key ideas -- (a) creating several random partitions of the
data and fitting separate models to each in parallel, and (b) an ensemble approach to refine the random effect
estimates. The ensemble is created by using several runs of random partitioning of the data and averaging over the random effect estimates from each run. This combination of divide and conquer coupled with ensembles provides a simple yet effective
procedure. Due to multi-modal nature of the posterior distribution, careful initialization that synchronizes factor estimates across partitions is important. 
The partitioning method employed also has an impact on performance; we provide a detailed study of these issues also.

To summarize, we make the following \textbf{contributions}. We provide a careful study of fitting regression
based bilinear random effect (BIRE) models to binary response data that are commonplace in many web applications. We show that
when modeling rare response, previously proposed methods can be improved by exact sampling of factors
through an adaptive rejection sampling procedure. We provide a modeling strategy that scales to massive datasets
in a Map-Reduce framework. Our method is based on a divide and conquer strategy coupled with an ensemble approach. We show that exact sampling of random effects and carefully handling identifiability issues can make a significant difference in model performance.
We compare our method with various baselines on benchmark data and illustrate impressive gains on a content optimization
problem on the Today module of Yahoo! front page.

\section{Personalized Item Recommendation on Yahoo! Front Page Today Module}
\begin{figure}[ht]
\begin{center}
\includegraphics[width=7cm]{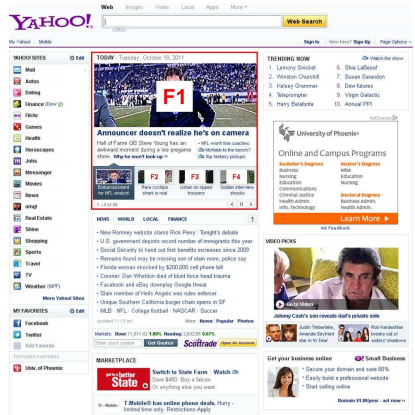}
\end{center}
\caption{A snapshot of the Yahoo! front page and its Today module.}
\label{figure::yahoo}
\end{figure}

The Yahoo! front page (\url{www.yahoo.com}) is a major web portal that attracts hundreds of millions of visitors every day. Figure \ref{figure::yahoo} shows a snapshot of the Yahoo! front page. The \em Today module \em is one of the most obvious modules on the web page. It displays article links on four positions labeled as F1 through F4.  The article link at the F1 position is displayed in a large and prime area in the module by default, while article links at F2 to F4 are displayed in the smaller footer area. A mouse hovering on a non-F1 article link will bring the article link to the prime area.  A click on an article link in the prime area will then lead the user to the actual article page. Since there are usually multiple links corresponding to one article, including title, related images and videos, we shall refer the entire bundle of the links as one article and treat a click on any of the links as a click on the article.
The personalized recommender system of the Today module is a combination of editorial oversight and statistical modeling. Statistical models are used to predict the probability of a user clicking an article; hence using the models to rank articles and show the best ones to the users gives optimal click through rates (CTR, the number of clicks divided by the number of views of the article) and improves user satisfaction and engagement for the front page. However, it has been well known that simply applying statistical modeling on a large and unscreened article inventory to optimize CTR is not optimal, sometimes even dangerous. For example, articles with salacious titles often have extremely high CTR, but in fact those are inappropriate for such a major web portal and it will cause Yahoo! to lose reputation. Therefore, in reality trained human editors at Yahoo! manually create and update the article inventory that contains 30-40 articles at any given time, and statistical CTR prediction models such as the ones described in this paper, pick the best 4 articles to show on position F1-F4 from the inventory based on the user's covariates and previous browsing history. Also, please note that the lifetime of each article in the inventory is usually quite short, ranging from several hours to 1 day.

\section{Regression-based Bilinear Random Effects Model}
\label{sec:model}

In this section, we describe a probabilistic bilinear random effects (BIRE)
model that leverages covariates to handle the cold-start problem and has
been shown to provide state-of-the-art performance on a number of
relatively small datasets~\citep{agarwal2009regression,gmf}.  We only
describe the model with logistic link function for binary response, and refer the readers
to~\cite{gmf} for the Gaussian model for numeric response and Poisson
model for count data.


\parahead{Notation}.  Let $y_{ij}$ denote whether user $i$ clicks item
$j$.  Since we always use $i$ to denote a user and $j$ to denote an
item, by slight abuse of notations, we let $x_{i}$, $x_j$ and $x_{ij}$
denote covariate vectors of user $i$, item $j$ and pair $(i,j)$. For
example, the user covariate vector $x_i$ may include age, gender and
behavioral covariates. The item covariate vector $x_j$ may include content
categories, keywords, named entities, etc.  The covariate vector
$x_{ij}$ contains observation-specific covariates that are not entirely
attributable to either user or item, e.g., time-of-day of the
observation, position of the item on the displayed web page.


\parahead{Model}
Our objective is to model the unobserved probability $p_{ij}$ that user $i$ would click item $j$.  Specifically, we assume a Bernoulli model using the logistic link function:
\begin{equation}
y_{ij} \sim \textrm{Bernoulli}(p_{ij}).
\end{equation}
Let $s_{ij} = \textstyle\log\frac{p_{ij}}{1-p_{ij}}$ denote the log odds. We model $s_{ij}$ by 
\begin{equation}
s_{ij} = f(x_{ij}) + \alpha_i + \beta_j + \bm{u}_i^\prime \bm{v}_j,
\end{equation}
where $f(x_{ij})$ is a regression function based on covariate vector
$x_{ij}$; $\alpha_i$ and $\bm{u}_i$ are latent random effects (factors) representing the user bias and the $r$-dimensional latent profile of user $i$, respectively; and $\beta_j$ and $\bm{v}_j$ are latent random effects (factors) representing the item popularity and the $r$-dimensional latent profile of item $j$.

\parahead{Flexible Regression Priors} 
Because the above model is
usually over-parameterized with a large number of latent factors, it is
important to regularize the factors to prevent over-fitting. A common
practice is to shrink the factors toward zero.  However, it fails to
handle the cold-start problem because the predicted factor values of
new users or items will all be zero.  A better approach is to shrink
factors to values predicted based on
features~\citep{agarwal2009regression,gmf}.  Specifically, we put the
following priors on $\alpha_i$, $\beta_j$, $\bm{u}_i$ and $\bm{v}_j$:
\begin{equation}
\begin{split}
\alpha_i \sim N(g(x_i), \sigma_{\alpha}^2), & ~~~~
\bm{u}_i \sim N(G(x_i), \sigma_u^{2} I), \\
\beta_j  \sim N(h(x_j), \sigma_{\beta}^2), & ~~~~
\bm{v}_j \sim N(H(x_j), \sigma_v^{2} I),
\end{split}
\end{equation}
where $g$ and $h$ are any choices of regression functions that return scalars, and $G$ and $H$ are regression functions that return $r$-dimensional
vectors.  These regression functions can be linear as in~\cite{agarwal2009regression} or non-linear (e.g., decision tree, forest, etc.) as in~\cite{gmf}.

To better understand the usefulness of regression priors, take $\bm{u}_i$ for example.  If user $i$ is a new user, then $\bm{u}_i$ is predicted by $G(x_i)$, where the regression function $G$ is learned based on users who interacted with some items in the training data. Let $G(x_i) = (G_1(x_i), ..., G_r(x_i))$.  One example of $G$ is to use a regression tree $G_k$ for each latent dimension $k$ to predict the value of the $k$-th dimension of a user's latent profile based on his/her covariate vector.  If covariates are predictive, we would be able to make accurate click rate prediction for new users.

\commentout{
\parahead{Model}. Our objective is to model the unknown latent variable $s_{ij}$ for user $i$ and item $j$ based on the observed response $y_{ij}$. In particular, for numeric ratings, it is common to assume a Gaussian model:
$$
y_{ij} \sim N(s_{ij}, \sigma^2).
$$ 

For binary responses (e.g., click or
no-click), it is common to assume a Bernoulli model using Logistic link function:
$$
y_{ij} \sim \textrm{Bernoulli}(p_{ij}) \mbox{ and let } s_{ij} = \textstyle\log\frac{p_{ij}}{1-p_{ij}}.
$$
For counts, we can assume a Poisson model:
$$
y_{ij} \sim \textrm{Poisson}(n_{ij}\,p_{ij}) \mbox{ and let } s_{ij} = \textstyle\log p_{ij},
$$
where $n_{ij}$ is a normalization constant associated with $y_{ij}$. 

We model $s_{ij}$ by 
$$
s_{ij} = f(x_{ij}) + \alpha_i + \beta_j + \bm{u}_i^\prime \bm{v}_j,
$$ 
where $f(x_{ij})$ is a regression function based on covariate vector
$x_{ij}$. $\alpha_i$ and $\bm{u}_i$ are main effects and $r$-dimensional factors from user $i$. $\beta_j$ and $\bm{v}_j$ are main effects and $r$-dimensional factors from item $j$.
}

\section{Model Fitting for Binary Data}
\label{sec::fitting}

In this section, we describe the model fitting procedure based on the
Monte Carlo Expectation Maximization (MCEM) algorithm for datasets
that can fit in a single machine. This is the 
building block for large data scenario stored in distributed clusters as described in
Section \ref{sec::parallel}.  We first describe the general fitting
procedure using the MCEM algorithm \citep{booth99mcem} in Section
\ref{subsec::MCEM} and then introduce two ways to handle binary
response in the E-Step: the variational approximation method
(\textbf{VAR}) in Section \ref{subsec::variational} and the adaptive
rejection sampling method (\textbf{ARS}) in Section \ref{subsec::ars}.
Finally, we briefly describe the M-Step in Section
\ref{subsec::Mstep}; it is the same as that in~\cite{gmf} but repeated
here for comprehensiveness.

\subsection{The MCEM Algorithm}
\label{subsec::MCEM}

Let $\bm{\Theta} = (f, g, h, G, H, \sigma_\alpha^2, \sigma_u^2,
\sigma_\beta^2, \sigma_v^2)$ be the set of prior parameters (also referred to
as hyper-parameters).  Let
$\bm{\Delta} = \{\alpha_i, \beta_j, \bm{u}_i, \bm{v}_j\}_{\forall i,j}$ be the set of latent random effects (also referred to as factors).  Let $\bm{y}$ denote the set of observed binary response.  For $M$ users and $N$ items, the {\em complete
data log-likelihood} is given by
\begin{equation}
\begin{split}
& \log L(\bm{\Theta}; \bm{\Delta}, \bm{y}) = \log \Pr[\bm{y}, \bm{\Delta} | \bm{\Theta}] = \mbox{ constant } \\
& ~~ - \textstyle \sum_{ij} y_{ij} \log(1+\exp(-f(x_{ij})-\alpha_i-\beta_j-\bm{u}_i'\bm{v}_j)) \\
& ~~ - \textstyle \sum_{ij} (1-y_{ij}) \log(1+\exp(f(x_{ij})+\alpha_i+\beta_j+\bm{u}_i'\bm{v}_j))\\
& ~~ - \textstyle \frac{1}{2 \sigma_{\alpha}^2} \sum_{i}
				(\alpha_i - g(x_i))^{2}
				- \frac{M}{2} \log \sigma_{\alpha}^2 \\
& ~~ - \textstyle \frac{1}{2 \sigma_{\beta}^2} \sum_{j}  
				 (\beta_j - h(x_j))^{2}
				- \frac{N}{2} \log \sigma_{\beta}^2  \\
& ~~ - \textstyle \frac{1}{2 \sigma_{u}^2} \sum_{i} 
				||\bm{u}_i - G(x_i)||^2
				- \frac{Mr}{2} \log \sigma_{u}^2 \\
& ~~ - \textstyle \frac{1}{2 \sigma_{v}^2} \sum_{j} 
				\|\bm{v}_j - H(x_j)\|^2
				- \frac{Nr}{2} \log \sigma_{v}^2. \\
\end{split}
\end{equation}
To apply the standard EM algorithm \citep{dempster77em}, we can treat $\bm{\Delta}$ as missing values and find the optimal estimate for $\bm{\Theta}$ that maximizes the marginal likelihood
\begin{equation}
\textstyle \Pr[\bm{y} | \bm{\Theta}] = \int L(\bm{\Theta}; \bm{\Delta}, \bm{y}) \,d\,\bm{\Delta}.
\end{equation}
The EM algorithm iterates between
an E-step and a M-step until convergence.  Let $\hat{\bm{\Theta}}^{(t)}$ denote the current estimated value of $\bm{\Theta}$ at the beginning of the $t$-th iteration.
\begin{itemize}
\item {\bf E-step:}  We take expectation of the complete data log likelihood with respect to the posterior distribution of the latent random effects $\bm{\Delta}$ conditional on
observed data $\bm{y}$ and the current estimate of $\bm{\Theta}$; i.e., compute
\begin{equation}
q_t(\bm{\Theta}) = E_{\bm{\Delta}}[\log L(\bm{\Theta}; \bm{\Delta}, \bm{y}) \,|\, \hat{\bm{\Theta}}^{(t)}, \bm{y}]
\end{equation}
as a function of $\bm{\Theta}$, where the expectation is taken over the posterior distribution of $p(\bm{\Delta} \,|\, \hat{\bm{\Theta}}^{(t)}, \bm{y})$ and $\hat{\bm{\Theta}}^{(t)}$ is treated as a set of constants.  The output of the E-step consists of a set of sufficient statistics to be used in the M-step.

\item {\bf M-step:} We maximize the expected complete data log-likelihood from the E-step to
obtain updated values of $\bm{\Theta}$; i.e., find 
\begin{equation}
\hat{\bm{\Theta}}^{(t+1)} = \arg\max_{\bm{\Theta}} ~ q_t(\bm{\Theta}).
\end{equation}
\end{itemize}
Note that in the E-Step, the posterior $p(\bm{\Delta} \,|\, \hat{\bm{\Theta}}^{(t)}, \bm{y})$ is not available in closed form.  Thus, we compute Monte Carlo means based on Gibbs samples following the MCEM algorithm \citep{booth99mcem,agarwal2009regression,gmf}.  According to \cite{ruslan:icml08b,agarwal2009regression}, this approach provides better predictive accuracy and avoids over-fitting while it remains scalable compared to other choices such as the iterative conditional mode (ICM) algorithm.

Before we describe the E-Step, we first provide the formula for $q_t(\bm{\Theta})$.  Let $\hat{\delta} = E[\delta | \hat{\bm{\Theta}}^{(t+1)}, \bm{y}]$ and $\hat{V}[\delta] = \textit{Var}[\delta | \hat{\bm{\Theta}}^{(t+1)}, \bm{y}]$, where $\delta$ can be one of $\alpha_i$, $\beta_j$, $\bm{u}_i$ and $\bm{v}_j$. Then, we have
\begin{equation}
\begin{split}
& q_t(\bm{\Theta}) = E_{\bm{\Delta}}[\log L(\bm{\Theta}; \bm{\Delta}, \bm{y}) \,|\, \hat{\bm{\Theta}}^{(t)}, \bm{y}] = \mbox{ constant } \\
& ~~ - \textstyle \sum_{ij} y_{ij} E[\log(1+\exp(-f(x_{ij})-\alpha_i-\beta_j-\bm{u}_i'\bm{v}_j))] \\
& ~~ - \textstyle \sum_{ij} (1-y_{ij}) E[\log(1+\exp(f(x_{ij})+\alpha_i+\beta_j+\bm{u}_i'\bm{v}_j))]\\
& ~~ - \textstyle \frac{1}{2 \sigma_{\alpha}^2} \sum_{i}
				\big((\hat{\alpha}_i - g(x_i))^{2} + \hat{V}[\alpha_i]\big)
				- \frac{M}{2} \log \sigma_{\alpha}^2 \\
& ~~ - \textstyle \frac{1}{2 \sigma_{\beta}^2} \sum_{j}  
				\big( (\hat{\beta}_j - h(x_j))^{2} + \hat{V}[\beta_j] \big)
				- \frac{N}{2} \log \sigma_{\beta}^2  \\
& ~~ - \textstyle \frac{1}{2 \sigma_{u}^2} \sum_{i} 
				\big( ||\hat{\bm{u}}_i - G(x_i)||^2 + \textrm{tr}(\hat{V}[\bm{u}_i]) \big)
				- \frac{Mr}{2} \log \sigma_{u}^2 \\
& ~~ - \textstyle \frac{1}{2 \sigma_{v}^2} \sum_{j} 
				\big( \|\hat{\bm{v}}_j - H(x_j)\|^2 + \textrm{tr}(\hat{V}[\bm{v}_j])  \big)
				- \frac{Nr}{2} \log \sigma_{v}^2. \\
\end{split}
\end{equation}
It is easy to see that the sufficient statistics for maximizing $q_t(\bm{\Theta})$ are $\hat{\alpha}_i$, $\hat{\beta}_j$, $\hat{\bm{u}}_i$, $\hat{\bm{v}}_j$ for all $i$ and $j$, as well as $\sum_i \hat{V}[\alpha_i]$, $\sum_j \hat{V}[\beta_j]$, $\sum_i \textrm{tr}(\hat{V}[\bm{u}_i])$ and $\sum_j \textrm{tr}(\hat{V}[\bm{v}_j])$.  This set of quantities is computed based on $L$ Gibbs samples and is the output of the E-step.  We note that the first two terms are difficult to expand and we will use plug-in estimates of $\alpha_i$, $\beta_j$, $\bm{u}_i$ and $\bm{v}_j$ to determine a near optimal solution for $f$ in the M-step.

\subsection{Variational Method in E-Step}
\label{subsec::variational}
Since $E_{\bm{\Delta}}[\log L(\bm{\Theta}; \bm{\Delta}, \bm{y}) \,|\, \hat{\bm{\Theta}}^{(t)}]$ is not available in closed form, we compute the Monte-Carlo 
expectation based on $L$ samples generated by a Gibbs sampler~\citep{gelfand95gibbssampling}. The Gibbs sampler repeats the following procedure $L$ times.  In the following text, we use $(\delta \,|\, \textrm{Rest})$, where $\delta$ can be one of $\alpha_i$, $\beta_j$, $\bm{u}_i$, and $\bm{v}_j$, to denote the conditional distribution of $\delta$ given all the other latent random effects and the observations $\bm{y}$.  Let $\mathcal{I}_j$ denote the set of users who rated item $j$, and $\mathcal{J}_i$ denote the set of items rated by user $i$.

The variational approximation is based on~\cite{jaakkola2000bayesian} and was proposed in~\cite{agarwal2009regression} to factorize binary matrices.  We note that there is a typo in the variational approximation formula in~\cite{agarwal2009regression}.  The basic idea is to transform binary response values into Gaussian response values before each EM iteration and then just use the E-Step and M-Step of the Gaussian model.

Let $\xi_{ij}$ be a parameter associated with each observed $y_{ij}$.  We can set all $\xi_{ij} = 1$ initially.
\begin{itemize}
\item Before each E-step, create pseudo Gaussian response for each binary observation $y_{ij} \in \{0,1\}$.  The pseudo Gaussian response is $r_{ij} = \frac{2 y_{ij} - 1}{4\lambda(\xi_{ij})}$ with variance $\sigma_{ij}^2 = \frac{1}{2\lambda(\xi_{ij})}$, where $\lambda(\xi) =
\frac{1}{4\xi}\tanh\left(\frac{\xi}{2}\right)$.

\item Run the E-step using Gaussian pseudo observations $(r_{ij}, \sigma_{ij}^2)$.  Details will be provided later.
\item Run the M-step in Section \ref{subsec::Mstep}.
\item After the M-step, for each observation $y_{ij}$, set $\xi_{ij} = \sqrt{E[s_{ij}^2]}$.
\end{itemize}

Now we describe the details of the E-step given the pseudo Gaussian observations $(r_{ij}, \sigma_{ij}^2)$.  Repeat the following steps $L$ times to draw $L$ samples of $\bm{\Delta}$.
\begin{itemize}
\item{Draw $\alpha_i$ from Gaussian posterior $p(\alpha_i|\mbox{Rest})$} for each user $i$.
{\small\begin{equation}
\begin{split}
\mbox{Let } o_{ij} & = r_{ij} - f(x_{ij}) - \beta_j 
	 - \bm{u}_{i}^{\prime} \bm{v}_j, \\
\textit{Var}[\alpha_i|\mbox{Rest}] & = \textstyle
	( \frac{1}{\sigma_{\alpha}^2} + 
		\sum_{j \in \mathcal{J}_i} \frac{1}{\sigma_{ij}^{2}} )^{-1}. \\
E[\alpha_i|\mbox{Rest}] & = \textstyle
	\textit{Var}[\alpha_i|\mbox{Rest}]
	( \frac{g(x_i)}{\sigma_{\alpha}^2} + 
		 \sum_{j \in \mathcal{J}_i} \frac{o_{ij}}{\sigma_{ij}^{2}} ).
\end{split}
\end{equation}}
\item Draw $\beta_j$ for each item $j$ (similar to above).
\item{Draw $\bm{u}_i$ from Gaussian posterior $(\bm{u}_i \,|\, \textrm{Rest})$} for each user $i$.
{\small\begin{equation}
\begin{split}
\mbox{Let } o_{ij} & =  r_{ij} - f(x_{ij}) - 
						\alpha_i - \beta_j,  \\
\textit{Var}[\bm{u}_i|\mbox{Rest}] & = \textstyle
	( \frac{1}{\sigma_u^2} I + 
	  \sum_{j \in \mathcal{J}_i} 
			\frac{\bm{v}_j \bm{v}_j^{\prime}}{\sigma_{ij}^{2}} 
	)^{-1}. \\
E[\bm{u}_i|\mbox{Rest}] & = \textstyle
	\textit{Var}[\bm{u}_i|\mbox{Rest}]
	( \frac{1}{\sigma_u^2} G(x_i) + 
		 \sum_{j \in \mathcal{J}_i} \frac{o_{ij} \bm{v}_j}{\sigma_{ij}^{2}} ).
\end{split}
\end{equation}}
\item Draw $\bm{v}_j$ for each item $j$ (similar to above).
\end{itemize}

\subsection{Adaptive Rejection Sampling in E-Step}
\label{subsec::ars}
Although for binary data and logistic link function the conditional
posterior $p(\alpha_i|\mbox{Rest})$, $p(\beta_j|\mbox{Rest})$,
$p(\bm{u}_i|\mbox{Rest})$ and $p(\bm{v}_j|\mbox{Rest})$ are not in
closed form, precise and efficient sampling from the posterior can still be achieved
through adaptive rejection sampling (ARS)~\citep{gilks1992derivative}. 
ARS is an efficient method
to draw samples from an arbitrary univariate density provided it is
log-concave. In our E-Step, we can draw a sample from the joint posterior distribution of
$\bm{\Delta}$ by drawing one number at a time sequentially from the univariate
posterior distribution of each individual random effect given all the others using Gibbs sampling.  To construct such a Gibbs sampler, we note that the univariate conditional
posterior distributions $p(\cdot|\mbox{Rest})$ are all log-concave;
hence ARS can be applied.

\begin{figure}[ht]
\vspace{-3in}
\begin{center}
\includegraphics[width=9cm]{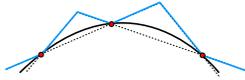}
\end{center}
\caption{Illustration of upper and lower bounds of an arbitrary (log) density function.}
\label{fig:ars-illustratation}
\end{figure}

In general, rejection sampling (RS) is a popular method used to sample
from a univariate distribution. Suppose we want to draw a sample from
a non-standard distribution with density $p(x)$. If one can find another
density $e(x)$ that is easier to sample from and approximates $p(x)$ well
and has tails heavier than $p(x)$, then $e(x)$ can be used to do rejection sampling. The key
is to find a constant $M$ such that $p(x) \leq M e(x)$ for all points $x$ such that $p(x) > 0$.
For example, the blue curve in Figure~\ref{fig:ars-illustratation} is $M e(x)$ and the black solid curve is $p(x)$.
The algorithm then is simple: We repeat the following steps until we obtain a valid sample. First we draw a number $x^{*}$ from $e(x)$. Then with probability $\frac{p(x^{*})}{Me(x^{*})}$, we accept $x^{*}$ as a valid sample; otherwise, we reject it.

Notice that $\frac{p(x^{*})}{Me(x^{*})}$ is always between 0 and 1.
This algorithm can be shown to provide a sample from
$p(x)$, and the acceptance probability is $1/M$.  Finding an $M$ that is small often involves knowing
the mode of $p(x)$; it is also important to find a good matching density $e(x)$ in practice.
ARS addresses both the issues. It finds a good matching density $e(x)$ that is composed of piecewise
exponentials; i.e., $\log e(x)$ is piecewise linear like the blue curve in Figure~\ref{fig:ars-illustratation}.
ARS does not need to know the mode of $p(x)$, and the only requirement is the log-concavity of $p(x)$, which is true for our problem.  The piecewise exponentials are constructed by creating an upper envelope
of the target log density. Further, the procedure is adaptive and uses the rejected points to further
refine the envelope which reduces the rejection probability for future samples.

\commentout{
construct envelope functions of the log of the target density, which
is then used in rejection-sampling. The envelope is updated to
correspond more closely to the log of the target density whenever a
sample is rejected. Hence as time goes subsequent proposed samples
need less and less number of evaluations of the target density (which
is usually the most expensive computational bottleneck). The
derivative-free ARS method is proposed by \cite{gilks1992derivative}
but requires the target density function to be
log-concave. \cite{gilks1995adaptive} generalized this approach to
density function that do not have log-concavity property. For our
problem since $p(\alpha_i|\mbox{Rest})$, $p(\beta_j|\mbox{Rest})$,
$p(\bm{u}_i|\mbox{Rest})$ and $p(\bm{v}_j|\mbox{Rest})$ are all
log-concave functions, we simply follow
\cite{gilks1992derivative}. Note that traditional methods such as
Metropolis-Hastings \cite{huelsenbeck2001mrbayes} can also be applied
to obtain posterior samples, however the proposals need to be chosen
very cleverly to reduce the rejection rate while ARS automatically
solves the problem.
}

We use the derivative-free ARS process from \cite{gilks1992derivative} which can be
briefly described as follows: Suppose we want to obtain a sample $x^*$
from a log-concave target density function $p(x)$.  We start from at
least 3 initial points such that at least
one point lies on each side of the mode of $p(x)$ (this is ensured by looking
at the derivative of the density, which does not require actual mode computation).
A lower bound $\textit{lower}(x)$ of $\log p(x)$ is constructed from the chords joining the evaluated
points of $p(x)$ with the vertical lines at the extreme points.  For example, the dotted piecewise linear curve in Figure~\ref{fig:ars-illustratation} is $\textit{lower}(x)$, while the solid black curve is $\log p(x)$.
An upper bound $\textit{upper}(x)$ is also constructed by extending the chords to their intersection points.  For example, the blue piecewise linear curve in Figure~\ref{fig:ars-illustratation} is $\textit{upper}(x)$.
The envelope function $e(x)$ (upper bound) and the squeezing function
$s(x)$ (lower bound) are created by exponentiating the piece-wise linear upper and
lower bounds of $\log p(x)$; i.e.,
$
e(x) = \exp(\textit{upper}(x))
~~\textrm{and}~~
s(x) = \exp(\textit{lower}(x)).
$
Let $e_1(x)$ be the corresponding density function derived from $e(x)$; i.e.,
$
e_1(x) = \frac{e(x)}{\int e(x) dx}.
$
The sampling produce works as follows: Repeat the following steps until we obtain a valid sample.
\begin{itemize}
\item Draw a number $x^*$ from $e_1(x)$ and another number $z \sim \mbox{Unif}(0,1)$, independently. 
\item If $z \leq \frac{s(x^*)}{e(x^*)}$, accept $x^*$ as a valid sample.
\item If $z\leq \frac{p(x^*)}{e(x^*)}$, accept $x^*$ as a valid sample; otherwise, reject $x^*$.
\item If $x^*$ is rejected, update $e(x)$ and $s(x)$ by constructing new chords using $x^*$.
\end{itemize}
This goes on iteratively until one sample is accepted. Note that using the squeezing function as the 
acceptance criteria implies partial information from the original density $p(x)$; 
Testing $x^*$ based on the squeezing function first is to save computation since the squeezing
function is readily available from the constructed envelope and evaluation of $p(x^*)$ is usually costly.

The ARS-based E-step works as follows: Repeat the following steps $L$ times to draw $L$ samples of $\bm{\Delta}$.
\begin{itemize}
\item Sample $\alpha_i$ from $p(\alpha_i|\mbox{Rest})$ for each user $i$ using ARS. The log of the target density is given by
{\small
\begin{equation}
\begin{split}
& \log p(\alpha_i|\mbox{Rest}) = \mbox{ constant } \\
& ~~ - \textstyle \sum_{j\in \mathcal{J}_i} y_{ij} \log(1+\exp(-f(x_{ij})-\alpha_i-\beta_j-\bm{u}_i'\bm{v}_j)) \\
& ~~ - \textstyle  \sum_{j\in \mathcal{J}_i} (1-y_{ij}) \log(1+\exp(f(x_{ij})+\alpha_i+\beta_j+\bm{u}_i'\bm{v}_j))\\
& ~~ - \textstyle \frac{1}{2 \sigma_{\alpha}^2}
				(\alpha_i - g(x_i))^{2}.\\
\end{split}
\end{equation}
}
\item Sample $\beta_j$ for each item $j$ (similar to above).

\item Sample $\bm{u}_i$ from $p(\bm{u}_i|\mbox{Rest})$ for each user $i$.
Since $\bm{u}_i$ is an $r$-dimensional vector, for each $k=1,\cdots, r$ we sample $u_{ik}$ from $p(u_{ik}|\mbox{Rest})$  using ARS. The log of the target density is given by
{\small
\begin{equation}
\begin{split}
& \log p(u_{ik}|\mbox{Rest}) = \mbox{ constant } \\
& ~~ - \textstyle \sum_{j\in \mathcal{J}_i} y_{ij} \log(1+\exp(-f(x_{ij})-\alpha_i-\beta_j-u_{ik}v_{jk}\\
& ~~~~~~~~~~~~~~~~~~~~~~~~~~~~~~~~~~~~~~~~ \textstyle -\sum_{l\neq k}u_{il} v_{jl})) \\
& ~~ - \textstyle  \sum_{j\in \mathcal{J}_i} (1-y_{ij}) 
\log(1+\exp(f(x_{ij})+\alpha_i+\beta_j+u_{ik}v_{jk} \\
& ~~~~~~~~~~~~~~~~~~~~~~~~~~~~~~~~~~~~~~~~ \textstyle +\sum_{l\neq k}u_{il}v_{jl})) \\
& ~~ - \textstyle \frac{1}{2 \sigma_{u}^2}
(u_{ik} - G_k(x_i))^{2}.\\
\end{split}
\end{equation}
}
\item Sample $\bm{v}_j$ for each item $j$ (similar to above).
\end{itemize}

\parahead{Initial points for ARS}
The rejection rate of ARS depends on the initial points and the target density function. To reduce the rejection rate, \cite{gilks1995adaptive} suggest using the envelope function from the previous iteration of the Gibbs sampler to construct 5th, 50th and 95th percentiles as the 3 starting points. We adopted this approach in our sampling and 
observed roughly 60\% reduction in rejection rates.

\parahead{Centering} We note that the model proposed in Section
\ref{sec:model} is not identifiable. For example, if we let
$\tilde{f}(x_{ij})=f(x_{ij})-\delta$ and
$\tilde{g}(x_i)=g(x_i)+\delta$ where $\delta$ can be any constant, the model using $\tilde{f}$ and
$\tilde{g}$ is essentially the same as the one using $f$ and $g$. To
help identify the model parameters, we put constraints on the random effect
values.  Specifically, we require $\sum_i \alpha_i = 0$, $\sum_j
\beta_j = 0$, $\sum_i \bm{u}_i = \bm{0}$ and $\sum_j \bm{v}_j =
\bm{0}$.  These constraints induce dependencies among user random effects and
item random effects.  Instead of dealing with these dependencies in sampling,
we simply enforce these constraints after sampling by subtracting the
sample mean; i.e., after sampling all the random effects, compute $\bar{\alpha} =
\sum_i \hat{\alpha}_i/M$ and set $\hat{\alpha}_i = \hat{\alpha}_i -
\bar{\alpha}$ for all $i$, and so on.  Here, $M$ is the number of
users and $\hat{\alpha}_i$ is the posterior sample mean of $\alpha_i$.

\subsection{M-Step}
\label{subsec::Mstep}
In the M-step, we find the parameter setting $\bm{\Theta}$ that maximizes the expectation computed in the E-step 
\begin{equation}
q_t(\bm{\Theta}) = E_{\bm{\Delta}}[\log L(\bm{\Theta}; \bm{\Delta}, \bm{y}) \,|\, \hat{\bm{\Theta}}^{(t)}].
\end{equation}
It can be easily seen that $(f, \sigma^2)$, $(g, \sigma_\alpha^2)$, $(h, \sigma_\beta^2)$, $(G, \sigma_u^2)$, and $(H, \sigma_v^2)$ can be optimized by separate regressions. Here we simply describe how to estimate $(G, \sigma_u^2)$ since everything else is quite similar. Recall that $\hat{u}_{ik}$ and $\hat{V}[u_{ik}]$ denote the posterior sample mean and variance of $u_{ik}$ computed based on the $L$ Gibbs samples obtained in the E-step. 
It is easy to see that
\begin{equation}
\arg\max_G q_t(\bm{\Theta}) = \arg\max_G \sum_i ||\hat{\bm{u}}_i - G(x_i)||^2.
\end{equation}
Note that $G$ is part of $\bm{\Theta}$, and finding the optimal $G$ is solving a least squares regression problem using $x_i$ as covariates to predict
multivariate response $\hat{\bm{u}}_{i}]$.  For univariate regression models, we consider $G(x_i) = (G_1(x_i), ..., G_r(x_i))$, where each $G_k(x_i)$ returns a scalar.  In this case, for each $k$, we find $G_k$ by solving a regression problem that uses $x_i$ as features to predict $\hat{u}_{ik}$.
Let $\textrm{RSS}$ denote
the total residual sum of squares.  Then, $\sigma_u^2 =
(\sum_{ik} \hat{V}[u_{ik}] + \textrm{RSS}) /
(rM)$, which is obtained by setting the derivative of $q_t(\bm{\Theta})$ with respect to $\sigma_u^2$ to zero.

We note that obtaining the optimal $f$ (i.e., $\arg\max_f q_t(\bm{\Theta})$) is actually difficult because of the expectation of the log of some combination of random effects.  Thus, we use plug-in estimates; i.e., solve a logistic regression problem that uses $x_{ij}$ as features to predict $y_{ij}$ with offset $\hat{\alpha}_i + \hat{\beta}_j + \hat{\bm{u}}_i^\prime \hat{\bm{v}_j}$.

\section{Parallelized Model Fitting for Large Data}
\label{sec::parallel}
In this section we consider fitting algorithms for
large data sets that reside in distributed clusters and cannot fit into memory of a single machine. For such scenarios, fitting algorithms described in Section
\ref{sec::fitting} do not work. We provide a fitting strategy in a 
the Map-Reduce framework \citep{dean2008mapreduce}. We first apply the ``divide and conquer"
approach to partition the data into small partitions, and then run
MCEM on each partition to obtain estimates of $\bm{\Theta}$. The final
estimate of $\bm{\Theta}$ are obtained by averaging over
estimates of $\bm{\Theta}$ from all the partitions. Finally, given
$\bm{\Theta}$ fixed, we do $n$ {\em ensemble runs}, i.e. re-partition
the data $n$ times using different random seeds, and for each
re-partitioning we only run E-Step jobs on all partitions and then
average the results from them to obtain the final estimate of $\bm{\Delta}$. This algorithm is described in Algorithm \ref{algo:rlfmensemble}.

\begin{algorithm}
\caption{Parallel BIRE Model Fitting}
\label{algo:rlfmensemble}
\begin{algorithmic}
\STATE Initialize $\bm{\Theta}$ and $\bm{\Delta}$.
\STATE Partition data into $m$ partitions using random seed $s_0$.
\FOR{each partition $\ell \in \{1, ..., m\}$ running in parallel}
\STATE Run MCEM algorithm for $K$ number of iterations using VAR or ARS to obtain $\hat{\bm{\Theta}}_\ell$, the estimates of $\bm{\Theta}$ for each partition $\ell$.
\ENDFOR
\STATE Let $\hat{\bm{\Theta}}=\frac{1}{m}\sum\limits_{\ell=1}^m \hat{\bm{\Theta}}_\ell.$
\FOR{$k=1$ to $n$ running in parallel}
\STATE Partition data into $m$ partitions using random seed $s_k$.
\FOR{each partition $\ell \in \{1, ..., m\}$ running in parallel}
\STATE Run E-Step-Only job given $\hat{\bm{\Theta}}$ and obtain the posterior sample mean $\hat{\bm{\Delta}}_{k\ell}$ for all users and items in partition $\ell$.
\ENDFOR
\ENDFOR
\STATE For each user $i$, average over all $\hat{\bm{\Delta}}_{k\ell}$ that contain user $i$ to obtain $\hat{\alpha}_i$ and $\hat{\bm{u}}_i$.
\STATE For each item $j$, average over all $\hat{\bm{\Delta}}_{k\ell}$ that contain item $j$ to obtain $\hat{\beta}_i$ and $\hat{\bm{v}}_i$.
\end{algorithmic}
\end{algorithm}

\parahead{Partitioning the data} Extensive experiments conducted by us showed
that model performance depends crucially on data partitioning strategy used in the Map-Reduce phase, especially when data is sparse. A naive way of
randomly partitioning observations may not give good predictive
accuracy. For applications such as content optimization
\citep{agarwal08nips}, the number of users are often much larger 
than the number of items. Also, the number of observations available per user
is small for a large fraction of users; a typical item tends to have a relatively
larger sample size. In such cases, we recommend
partitioning the data by users, which guarantees that all data from a user belongs to the same partition, so that good user
random effects can be obtained. Similarly, when the number of items is larger
than the number of users, we recommend partitioning the data by items.
An intuitive explanation of this can be gleaned by looking at the conditional variance
of user random effect $\bm{u}_i$ using variational approximation given as 
$\textit{Var}[\bm{u}_i|\mbox{Rest}] =( \frac{1}{\sigma_u^2} I + 
	  \sum_{j \in \mathcal{J}_i} 
			\frac{\bm{v}_j \bm{v}_j^{\prime}}{\sigma_{ij}^{2}} )^{-1}.$
Assuming item random effects are known for the moment (or estimated with high
precision), if the user data is split into several partitions, the average
information gain (inverse variance) from the partitioned data is the
harmonic mean of information gain from individual partitions. The
information gain from the non-partitioned data can be written as the
arithmetic mean of the individual information gains. Since harmonic
mean is less than arithmetic mean, the information loss in estimating the user
random effects by partitioning is the difference
in arithmetic and harmonic means. When the information in partitions
becomes weak, this gap increases. Hence, with sparse user data, it
is prudent to partition by users.

\parahead{Estimates of $\bm{\Theta}$} We note the $\bm{\Theta}$ estimate obtained from
each random partition is unbiased,
fitting a model on each partition and then averaging the M-step
parameters $\hat{\bm{\Theta}}_\ell$ for $\ell=1,\cdots,m$ provide 
an estimate that is still unbiased and has lower variance due to
lack of positive correlations among estimates. The correlations are absent due to
the random partitioning. 
Before running the MCEM algorithm, the initial values of $\bm{\Theta}$ for all partitions are the same.  In particular, we start with zero-mean priors; i.e., $g(x_i) = h(x_j) = 0$ and $G(x_i) = H(x_j) = \bm{0}$.
To improve parameter estimation, one may synchronize the parameters among partitions and run another round of MCEM iterations; i.e., one may re-partition the data and use the obtained $\hat{\bm{\Theta}}$ as the initial values of $\bm{\Theta}$ to run another round of MCEM iterations for each partition to obtain a new estimate of $\bm{\Theta}$.
However, we observe in practice that iteratively running this process
does not give significantly better predictive accuracy,
but instead adds complexity and training time.

\parahead{Estimates of $\bm{\Delta}$} For each run in the ensemble, it
is essential to use a different random seed for partitioning the data,
so that the mix of users and items in partitions across different runs
would be different. Given $\hat{\bm{\Theta}}$, for each run in the
ensemble, we only need to run E-step once for each partition and
obtain the final user and item random effects by taking the average.
Again, the random partitioning ensures uncorrelated estimates from members of the ensemble
and leads to variance reduction through averaging.

\parahead{More identifiability issues} 
After centering the model
is in fact still non-identifiable because of two reasons: (a). Since
$\bm{u}_i'\bm{v}_j=(-\bm{u}_i)'(-\bm{v}_j)$, switching signs of
$\bm{u}$ and $\bm{v}$ (also the corresponding cold-start parameters)
does not change the log-likelihood. (b). For any two random effect indices $k$ and $l$, switching $u_{ik}$ with $u_{il}$,
$v_{jk}$ with $v_{jl}$ for all users and items simultaneously also would not change the
log-likelihood, given that the corresponding cold-start parameters are
also switched. We have found empirically that both of the
identifiability issues do not matter for small data sets,
especially single-machine runs. However, for large data sets such as
the Yahoo! frontpage data and $G$, $H$ defined as linear regression
function, we observe that for each partition after the MCEM step we
obtain significantly different fitted values of $G$ and $H$, so that after
averaging over all the partitions the resulting coefficient matrices
for $G$ and $H$ become almost zero. Hence the identifiability issue can
become severe while fitting parallelized BIRE models for
large data sets.

\parahead{Solution to the identifiability issues}
For (a), we put constraints on the item random effects $\bm{v}$ so that it is always positive. This can be done through simply putting a sampling lower bound (i.e. always sample positive numbers) in the adaptive rejection sampling. Note that after using this approach we do not need to do centering on $\bm{v}$ any more. For (b), we first let $\sigma_v^2=1$ and change the prior of $\bm{u}_i$ from $N(G(x_i),\sigma_u^2 I)$ to $N(G(x_i),\Sigma_u)$, where $\Sigma_u$ is a diagonal variance matrix with diagonal values $\sigma_{u1}\geq \sigma_{u2}\geq\cdots\geq \sigma_{ur}$. The model fitting is very similar; but after each M-step we re-sort all the random effects by the fitted $\sigma_{uk}$'s for $k=1,\cdots,r$ to satisfy the constraint.

\section{Experiments}

We evaluate the proposed methods to address two main questions: (1)
How do different techniques for handling binary response compare?  (2)
How do different methods perform in a real, large-scale web recommender
system?  For the first question, we compare variational approximation,
adaptive rejection sampling and stochastic gradient descent on
balanced and imbalanced binary datasets created from the public
MovieLens 1M dataset.  For the second question, we first evaluate the
predictive performance using a small balanced data set with
heavy users of the Today module on the Yahoo! front page to allow
comparison in the single-machine fitting scenario, and then provide
complete end-to-end evaluation in terms of the click-lift metric through a
recently proposed unbiased offline evaluation
method (\cite{li2011unbiased}, which has been shown to be able to
approximate the online performance) based on massive imbalanced binary response data
collected from the Today module.

\parahead{Methods}
We consider the following different models or fitting methods, all used with 10 factors per user/item throughout the experiments:
\myItemizeBegin
\item {\bf \FEAT} is the covariate-interaction-only model which serves as our baseline.  Specifically, the model is
$$s_{ij} = f(x_{ij}) + g(x_i) + h(x_j) + G(x_i)^\prime H(x_j),$$
where $g$, $h$, $G$ and $H$ are unknown regression functions, fitted by the standard conjugate gradient descent method on each partition and averaging over estimates from all partitions to obtain estimates of $g$, $h$, $G$ and $H$; no ensemble run is needed.
\item {\bf \VAR} is our regression-based BIRE model fitted by variational approximation in the MCEM algorithm.
\item {\bf \ARS} is our regression-based BIRE model fitted by centered adaptive rejection sampling algorithm in each E-step of the MCEM algorithm.
\item {\bf \ARSID} is our regression-based BIRE model fitted by centered adaptive rejection sampling algorithm in each E-step of the MCEM algorithm, incorporating positive constraints on the item random effect (factor) $\bm{v}$ and ordered diagonal prior covariance matrix of $\bm{u}$ (see Section \ref{sec::parallel} for more details).
\item {\bf SGD} is a method that fits a similar BIRE model using stochastic gradient descent.  We obtained the code from~\cite{markus}.  Specifically, the model is
$$
s_{ij} = (\alpha_i+\bm{u}_i+ \bm{U} \bm{x}_i)'(\beta_j+\bm{v}_j +  \bm{V} \bm{x}_j),
$$
where $\bm{U}$ and $\bm{V}$ are unknown coefficient matrices for cold-start to map the covariate vectors $\bm{x}_i$ and $\bm{x}_j$ into the $r$-dimensional latent space.  For binary response with logistic link function, it minimizes the following loss function
{\small \begin{eqnarray*}
\sum_{ij} y_{ij} \log(1+\exp(-s_{ij}))+\sum_{ij}(1-y_{ij})\log(1+\exp(s_{ij}))\\
+\lambda_u\sum_i\norm{\bm{u}_i}^2+\lambda_v\sum_j\norm{\bm{v}_j}^2 + \lambda_U\norm{\bm{U}}^2+\lambda_V\norm{\bm{V}}^2,
\end{eqnarray*}}
where $\lambda_u$, $\lambda_v$, $\lambda_U$ and $\lambda_V$ are tuning parameters, and $\norm{\bm{U}}$ and $\norm{\bm{V}}$ are Frobenius norms.  Since this code has not been parallelized, we only use it in experiments on small datasets.  Trying different tuning parameter
values can be computationally expensive. 
In the experiments, we set $\lambda_u=\lambda_v=\lambda_U=\lambda_V=\lambda$
with $\lambda$ varying from 0, $10^{-6}$, $10^{-5}$, $10^{-4}$ and
$10^{-3}$.  We also tuned the learning rate by trying $10^{-5}$,
$10^{-4}$, $10^{-3}$, $10^{-2}$ and $10^{-1}$.
\myItemizeEnd
In \FEAT, \VAR, \ARS and MCEM-ARSID, we use linear regression functions for $g$, $h$, $G$ and $H$. 

\commentout{
\subsection{Algorithm Setup}
When running \VAR, \ARS and \ARSID for small data sets (e.g. The MovieLens 1M data and the small Yahoo! frontpage data), we ran the MCEM algorithm for 30 iterations, and for each iteration we collected 200 Gibbs samples in the E-step with 2 extra burn-ins.

When we ran parallel algorithms on large data sets, to obtain
estimates of $\bm{\Theta}$, we ran the MCEM algorithm for 15
iterations. For iteration number 1-5, 6-10 and 11-15, the
E-Step was done by collecting 13, 20 and 100 posterior samples
respectively using Gibbs sampling with 2 extra burn-ins. After we
obtained $\hat{\bm{\Theta}}$, we executed the E-step-only job in each
ensemble run. For each E-step-only job we also collected 200 posterior
samples using Gibbs sampling with 2 extra burn-ins.  Although we could
have used 200 Gibbs samples in the E-step of every EM iteration, we
chose to use a small number of samples in early iterations, because
this would be more computationally efficient while not losing
predictive accuracy: In the initial stages, M-step estimates are far from optimal and we do not need too many
samples in the E-step to make a large move in the posterior space. In later iterations, when the parameters are closer to optimal,
larger numbers of samples are needed to control the sampling error, in order to move in the direction of the optimal solution.
}

\subsection{MovieLens 1M Data}
\label{sec:movielense}

We first compare three techniques for fitting BIRE-style models with binary response (\VAR, \ARS and SGD) on the benchmark MovieLens 1M dataset. Note that in \cite{agarwal2009regression}, the regression-based BIRE model has been proved to be significantly better than various baseline models, such as zero-mean BIRE model and the \em Filterbot \em from \cite{park06}.

\parahead{Data}
The MovieLens 1M data consists of 1M ratings with scare from 1 to 5 provided by 6,040 users on set of 3,706 movies. 
We create training-test split based on
the timestamps of the ratings; the first 75\% of ratings serve as training data and the rest
25\% as test data.  This split introduces many new users
(i.e. cold-start) in test data. 
To study how different techniques handle binary response with different degree of sparsity of the positive response, 
we consider two different ways of creating binary response: (1) An
imbalanced dataset is created by setting the response value to 1 if and only if the original 5-point rating value is 1; otherwise it is set to 0. The percentage of positive response in this dataset is around
5\%. (2) A balanced dataset is created by setting the response to 1 if the original rating
is 1, 2, or 3; otherwise it is set to 0. The percentage of positive response in this dataset is around 44\%.  We report the predictive performance of SGD, \VAR and \ARS in terms of the Area Under the ROC Curve (AUC) for both datasets in Table~\ref{table:MovieLens}.

\begin{table}
\label{table:MovieLens}
\begin{center}
\begin{tabular}{cccc}
\hline
& & \multicolumn{2}{c}{AUC} \\
Method & \# Partitions & Imbalanced & Balanced\\
\hline
SGD      &  1 & 0.8090 & 0.7413\\
MCEM-VAR &  1 & 0.8138 & 0.7576\\
MCEM-ARS &  1 & 0.8195 & 0.7563\\
\hline
         &  2 & 0.7614 & 0.7599\\
MCEM-VAR &  5 & 0.7191 & 0.7538\\
         & 15 & 0.6584 & 0.7421\\
\hline
         &  2 & 0.8194 & 0.7622\\
MCEM-ARS &  5 & 0.7971 & 0.7597\\
         & 15 & 0.7775 & 0.7493\\
\hline
\end{tabular}
\end{center}
\caption{AUC of different methods on the imbalanced and balanced MovieLens datasets (\#partitions$=1$ indicates single-machine runs)
}
\end{table}
\parahead{Comparison between \ARS and \VAR}
As can be seen from the Table~\ref{table:MovieLens}, \ARS and \VAR have similar performance and both slightly outperform SGD when running on a single machine (i.e., \#partitions = 1).
It is interesting to see that, when running on multiple machines with 2 to 15 partitions, \ARS and \VAR still have similar performance on the balanced
dataset, while on the imbalanced dataset \VAR becomes much worse 
when the number of partitions increases (causing more severe data sparsity). 
We note that the degradation of performance when the number of partitions increases is expected because, with more partitions, each partition would have less and sparser data, which leads to a less accurate model for the
partition.

\parahead{Comparison with SGD}
Since SGD is a popular fitting
method for SVD-style matrix factorization \citep{koren2009matrix}, we also discuss how our
sampling-based methods compare to SGD.
To obtain good performance for SGD, one has to try a large number of
different values of the tuning parameters and learning rates, while
our methods does not need such tuning because all the hyper-parameters are
obtained through the EM algorithm.  Trying different tuning parameter
values can be computationally expensive, and it is less efficient in exploring the parameter space compared to an EM algorithm. 
After our best-effort tuning using the test data, for imbalanced data, SGD achieves best performance 0.8090 with $\lambda=10^{-3}$ and learning rate $ = 10^{-2}$. For
balanced data, SGD achieves best performance 0.7413 with
$\lambda=10^{-6}$ and learning rate $ = 10^{-3}$.  Even tuning SGD on
test data, the best AUC numbers of SGD on both balanced and imbalanced datasets are still slightly worse than
the those of \VAR and \ARS (which did not touch test data before testing).

\commentout{
\begin{table}
\caption{AUC of different methods on the balanced MovieLens data set (\#partitions$=1$ indicates single-machine runs) 
}
\label{table:balancedMovieLens}
\begin{center}
\begin{tabular}{ccc}
\hline
Method & \# Partitions & AUC\\
\hline
SGD & 1 &  0.7413 \\
MCEM-VAR & 1 &  0.7576 \\
MCEM-ARS & 1 &  0.7563 \\
\hline
& 2 & 0.7599 \\
MCEM-VAR & 5 & 0.7538 \\
 & 15 &  0.7421 \\
\hline
& 2 & 0.7622 \\
MCEM-ARS & 5 & 0.7597 \\
 & 15 &  0.7493 \\
\hline
\end{tabular}
\end{center}
\end{table}
}


\subsection{Small Yahoo! Front Page Data}
\label{sec:exp-small}

\label{subsec::smallexpt}
\begin{table}
\label{table:fpkdd09}
\begin{center}
\begin{tabular}{ccccc}
\hline
Method & \# Partitions & Partition Method & AUC\\
\hline
FEAT-ONLY & 1 & -- & 0.6781 \\
SGD & 1 & -- & 0.7252 \\
MCEM-VAR & 1 & -- & 0.7374 \\
MCEM-ARS & 1 & -- & 0.7364 \\
MCEM-ARSID & 1 & -- & 0.7283 \\
\hline
& 2 & User & 0.7280 \\
MCEM-ARS & 5 & User & 0.7227 \\
 & 15 & User & 0.7178 \\
\hline
& 2 & User & 0.7294 \\
& 5 & User & 0.7172 \\
MCEM-ARSID & 15 & User & 0.7133 \\
& 15 & Event & 0.6924 \\
& 15 & Item & 0.6917 \\
\hline
\end{tabular}
\end{center}
\caption{AUC of different methods on the small Yahoo! front page dataset (\#partitions$=1$ indicates single-machine runs)
}
\end{table}

We now evaluate different BIRE model fitting methods on a previously analyzed Yahoo! front page dataset \citep{agarwal2009regression}, which allows comparison of these methods to prior work.

\parahead{Data} This dataset consists of 1.9M binary response values (click or non-click) 
obtained from about 30K heavy users interacting with 4,316 news
articles published in the Today module on the Yahoo! front page.  The
observations were sorted by their timestamps and the first 75\% of
them are used as training data and the rest 25\% as test data. The set
of user covariates include age, gender, geo-location and browsing
behavior that is inferred based on users' network-wide activity
(e.g. search, ad-clicks, page views, subscriptions etc.) Since the
original set of user covariates is large, dimension reduction was done
through principal component analysis ~\citep{agarwal2009regression},
and finally we obtained around 100 numerical user covariates. Item
covariates consist of 43 hand-labeled editorial categories.

Recall that the Today module displays article links on four positions labeled as
F1 through F4, where the F1 article resides in a large and prime area.
Also, a hover on a non-F1 article link will bring the article link to the prime area.  A click on an article link in the prime area will then lead the user to the actual article page.  For this data set, clicks on article links in the prime area are interpreted as positive response, while displays of article links in the prime area (i.e. hover on a non-F1 article) without subsequent clicks are considered as negative response. Also note that user visits with no click on any position were ignored.  
Hence, in this data set, the percentage of positive response is
close to 50\% --- it is a balanced data set.

\commentout{
Each position consists of a footer link, there is 
one story link that displays an expanded view of the footer links.
By default, the footer link on F1 is the story link. In this data,
user visits with no action on Today Module were ignored, only those
with a click on one of F1-F4 were considered. A click on F2-F3 was
construed as a negative for article link on F1, a click on F1 is interpreted as a positive.

The binary response of this data is created in the following way.
First note that, when Yahoo! Frontpage is displayed, one item (called
{\em default item}) is selected and displayed at the prime position
(called F1 position) in the Today module, and a user can change the
item at the F1 position by clicking some icons in the module to browse
through all the items in the content pool.  Notice that the fact that
a user did not click on the default item does not mean that he/she is
not interested in the item --- he/she may come to this page for a
different reason (e.g., search or check email, etc.).  Thus, when
creating this data set, we only included users who viewed non-default
items in the module.  If a non-default is displayed to a user, the
user must have interacted with the Today module.  In this case, if the
user clicks on the item, we generate positive response 1; otherwise we
generate negative response 0.  The display and click events of the
default item for any user visit are ignored, in order to cut down
potential noise.
}  

\parahead{Single-machine results}
We first discuss the AUC performance for \FEAT, \VAR, \ARS and \ARSID running on a single machine (i.e. 1 partition), shown in Table~\ref{table:fpkdd09}.  We observe that
\VAR, \ARS, \ARSID and SGD all
outperform \FEAT significantly. This is because these models
allow warm-start user random effects (those users having data in the
training period) to deviate from purely covariate-based predictions, in
order to better fit the data. On the other hand, since the test data
consists of many new users and new items, handling cold-start
scenarios is still important. It has been shown in~\cite{agarwal2009regression} that, for this data set, \VAR significantly improves upon BIRE models that use zero mean priors for random effects, which is
commonly applied in many recommender system problems such as Netflix, and \VAR also significantly outperformed various other collaborative filtering algorithms.
We note that the performance of \VAR, \ARS and \ARSID are all close. This suggests that for balanced datasets, different fitting methods for logistic models are similar. 
We also note that \ARSID perform slightly worse than \ARS, because adding constraints on the item random effects $\bm{v}$ reduces the flexibility of \ARSID.  We defer the discussion on when \ARSID can provide significant benefit to Section~\ref{sec:large-y-fp-data}.

\parahead{Comparison with SGD} 
Similar to what we see in Section~\ref{sec:movielense}, even with SGD tuned on
test data, the best AUC 0.7252 (achieved by using $\lambda=10^{-6}$
and learning rate $= 10^{-3}$) is still slightly worse than
the AUC of \VAR, \ARS and \ARSID for single machine runs.

\commentout{
Since SGD is a popular fitting
method for matrix factorization \cite{koren2009matrix}, we compare our
sampling-based methods to an SGD method provided by~\cite{markus}.
To obtain good performance for SGD, one has to try a large number of
different values of the tuning parameters and learning rates, while
our methods does not need such tuning (all the hyper-parameters are
obtained through the EM algorithm). Trying different tuning parameter
values can be quite computationally expensive, and it is less efficient in exploring
the parameter space compared to an EM algorithm. 
In our experiments, we ran SGD with 10 factors on a
single machine for 30 passes over the entire data.  To reduce tuning
effort, we assumed $\lambda_u=\lambda_v=\lambda_U=\lambda_V=\lambda$
with $\lambda$ varying from 0, $10^{-6}$, $10^{-5}$, $10^{-4}$ and
$10^{-3}$.  We also tuned the learning rate by trying $10^{-5}$,
$10^{-4}$, $10^{-3}$, $10^{-2}$ and $10^{-1}$.  After trying all
possible combinations of these $\lambda$ values with these learning
rates, we found that the test AUC ranges from 0.6457 to 0.7252, where
the best performance 0.7252 is achieved by letting $\lambda=10^{-6}$
and learning rate $10^{-3}$.  We note that, 
}

\parahead{Number of partitions}
In Table \ref{table:fpkdd09} for \ARS and \ARSID (10 ensemble runs for both), as the number of partitions grows, we observe the expected degradation of performance, similar to what we observed in Section \ref{sec:movielense}. However, even with 15 partitions on
such a small data set \ARS and \ARSID (user-based partitioning) still
significantly outperforms \FEAT. In general, increasing the number of partitions would
increase computational efficiency, but usually leads to worse
performance. We have observed in our experiments that for large data sets the computation time of $2N$ partitions is roughly half of using $N$ partitions. Therefore 
we use as few partitions as possible given our computational budget.

\parahead{Different partition methods}
In Table \ref{table:fpkdd09} we also show the performance of our
parallel algorithm \ARSID (10 ensemble runs) with different numbers of
partitions and various partition methods. As mentioned in Section
\ref{sec::parallel}, we note that partitioning the data by users
is better than event-based or item-based partitioning in our application.
The reason for this is that
in our application, there are generally more users than items in the
data; hence user partitions are less sparse. 

\subsection{Large Yahoo! Front Page Data}
\label{sec:large-y-fp-data}

In this subsection, we show the performance of our
parallel algorithms on a large Yahoo! front page dataset
where single-machine fitting algorithms are not
feasible.  An unbiased evaluation method is used to estimate
the expected \em click-lifts \em if these algorithms were used in 
the production system \citep{li2011unbiased}. \cite{agarwal2011modeling} used the same click-lift metric to measure the model performances.

\parahead{Data} The training data was collected from the Today module
on Yahoo! front page during June 2011, while the test events were
collected during July 2011. The training data includes all page views
by users with at least 10 clicks in the Today module, and consists of
8M users, $\sim$4.3K items and 1 billion binary observations.  To
remove selection bias in evaluating our algorithms, the test data is
collected from a randomly chosen user population where, for each user
visit, an article is selected at random from the content pool and
displayed at the F1 position.  We shall refer to this as
\textit{random bucket}, which consists of around 2.4M clicks with old
users who were seen in the training period as well as new ones.

Each user is associated with 124 behavior covariates that
reflect various kinds of user activities on the entire Yahoo! network.  Each item is associated with 43 editorial hand-labeled categories.
A click on an F1 article link is a positive observation, while a view of an F1 article link without a subsequent click is a negative observation.
The percentage of positive response here is much lower than that of
the small dataset (4\%-10\%) --- the increased
sparsity and imbalance introduces additional challenges.

\parahead{Experimental setup} Because article lifetimes in the Today
module are short (6-24 hours), almost all items in the test period are
new.  To provide good performance for new items, one may frequently
re-train BIRE models in the test period. However, since the
amount of data is large, frequent re-training is not a feasible
solution.  Note that the set of users that come to Yahoo! are much less dynamic than items, hence a viable solution is to assume the user random effects (factors) from the training period is fixed and learn the item random effects in an online fashion. More precisely, let $\bm{x}_{i}$ denote the
behavior covariate vectors of user $i$ and $\bm{u}_{i}$ denote the user random effect vector produced by a
BIRE model that is learned in training period. For each item $j$ at time $t$ in test period, we fit an individual online logistic regression (OLR) model as described in \cite{agarwal2010fast} with log-odds $\bm{x}^{'}_{i}\bm{\beta}_{jt} +
\bm{u}^{'}_{i}\bm{\delta}_{jt}$, where the unknown parameters
$(\bm{\beta}_{jt}, \bm{\delta}_{jt})$ are updated online after each test
epoch as we collect more data on each item. The OLR models are initialized 
with a prior $(\bm{\beta}_{j0},\bm{\delta}_{j0}) \sim
MVN(\bm{0},\sigma^{2}I).$ Notice that different BIRE fitting methods
generate different $\bm{u}_{i}$s. The performance of a method is based on
click-lift of the recommendations generated based on the OLR models
using the $\bm{u}_{i}$s.

\parahead{Unbiased evaluation} 
The goal of this set of experiments is to maximize total number of clicks.  The precision@1 metric computed on the random bucket test set was shown to provide an unbiased
measure of an algorithm's performance when it is actually implemented in production~\citep{li2011unbiased}. 
We provide a brief description of the evaluation metric below.

For an epoch $t$ (5-minute interval) in the test period, we do
the following:
\begin{enumerate}
\item Compute the predicted CTR of all articles in the pool for each event in epoch $t$ under the model, based on the user covariates and latent random effects. The estimates can use 
    all data before epoch $t$.
\item For each click event at time $t$ in the test data, we select the
    an article $j^{*}$ from the current pool with the highest predicted probability, 
     If the article that was actually clicked in the test data matches $j^{*}$, we give the model a reward;
    otherwise, we ignore it.
\end{enumerate}
At the end, we compute click lift metric based on the total reward received from the model. 
Mathematically, for the test data from a random bucket and model $M$, the score $S(M)$ can be defined as
\begin{equation}
\label{eqn:F1hit}
S(M) = \sum_{visits\mbox{ } with\mbox{ } click}1(\mbox{item clicked} = \mbox{item selected by M}).
\end{equation}

It has been proved that $S(M)$ is an
unbiased metric compared to the real click lift seen in a production system~\citep{li2011unbiased}. Because each article in the random bucket
has an equal probability to be displayed to users, the number of
matched view events for any model is expected to be the
same.  A better model to optimize CTR can match more click
events. For large amounts of data as in our case, the variance
of the click-lift metric for any model is very small; all
differences reported in our experiments have small p-values and are
statistically significant due to large sample size in the test data.

\parahead{Two baseline methods}.  To show that random-effect-based user covariates (i.e. $\bm{u}_i$) provide state-of-the-art performance to personalize content on Yahoo!
front page, in this paper we implement two baseline methods of generating user covariates based on users' past interaction on the Today module:

\textbf{ITEM-PROFILE}: Using training data we pick top 1000 items that
have highest number of views. We construct 1000-dimensional binary
user profiles to indicate whether in the training period this user has
ever clicked on this item (1 is clicked and 0 is non-clicked). For
cold-start users that did not show up in training data, we simply let
the binary profile vector to be all $0$.

\textbf{CATEGORY-PROFILE}: Since in this dataset each item
has 43 binary covariates indicating content categories to which the item belongs,
we build user-category preference profiles through the following
approach: For user $i$ and category $k$, denote the number of observed
views as $v_{ik}$ and number of clicks as $c_{ik}$. From the training
data we can obtain the global per-category CTR, denoted as $\gamma_k.$ We then
model $c_{ik}$ as $c_{ik}\sim \textit{Poisson}(v_{ik}\gamma_k\lambda_{ik})$,
where $\lambda_{ik}$ is the unknown user-category preference parameter. We assume $\lambda_{ik}$ has a Gamma prior $\textit{Gamma}(a,a)$, hence the posterior of $\lambda_{ik}$ becomes
$(\lambda_{ik}|v_{ik},c_{ik})\sim \textit{Gamma}(c_{ik}+a,v_{ik}\gamma_k+a).$
We use the log of the posterior mean, i.e. $\log(\frac{c_{ik}+a}{v_{ik}\gamma_k+a})$ as the profile covariate value for user $i$ on category $k$. Note that if we do not observe any data for user $i$ and category $k$, the covariate value becomes 0. $a$ is a tuning prior sample size parameter and can be obtained through cross-validation. By trying $a=$1, 5, 10, 15 and 20, we have found that for this data set $a=10$ is the optimal value.

\begin{table}
\label{table:ctrlift}
\begin{center}
\begin{tabular}{ccccc}
\hline
Method & \#Ensembled & Overall & Warm & Cold\\
& Runs & & Start & Start\\
\hline
ITEM-PROFILE & -- & 3.0\% & 14.1\% & -1.6\%\\
CATEGORY-PROFILE & -- & 6.0\% & 20.0\% & 0.3\% \\
MCEM-VAR & 10 & 5.6\% & 18.7\% & 0.2\% \\
MCEM-ARS & 10 & 7.4\% & 26.8\% & -0.5\%\\
MCEM-ARSID & 1 & 9.1\% & 24.6\% & 2.8\%\\
MCEM-ARSID & 10 & 9.7\% & 26.3\% & 2.9\% \\
\hline
\end{tabular}
\end{center}
\caption{The overall click lift over the user behavior covariate (BT) only model.}
\end{table}

\parahead{Experimental Results}
We evaluate all methods by reporting click-lift obtained through the unbiased evaluation method relative to
an online logistic model that only uses behavioral (BT) covariates $\bm{x}_{i}$; such a model does not incorporate users' past interaction with items --- its performance on heavy users has large room for improvement. In Table~\ref{table:ctrlift}, we summarize
the overall lift, warm start lifts (users seen in the training set), and cold-start lifts (new users).
All models produce lifts but the performance of MCEM-ARSID is the best for overall and cold-starts, and MCEM-ARS is the best for warm-starts. The reason that we see no lift for cold-start users on MCEM-ARS is because of the identifiability issues addressed in Section \ref{sec::parallel}. Although imposing positive constraints on the item random effects leads \ARSID to have slightly inferior performance than \ARS for warm-start users, it solves the identifiability issues quite well and hence gives the best performance for the cold-start users.
It is also interesting to see that \VAR is worse than CATEGORY-PROFILE, especially for warm-starts. We also observe that using the
ensemble trick improves results as evident from comparing MCEM-ARSID with 1 and 10 ensemble runs.

\begin{figure}[ht]
\begin{center}
\centerline{\includegraphics[width=7cm]{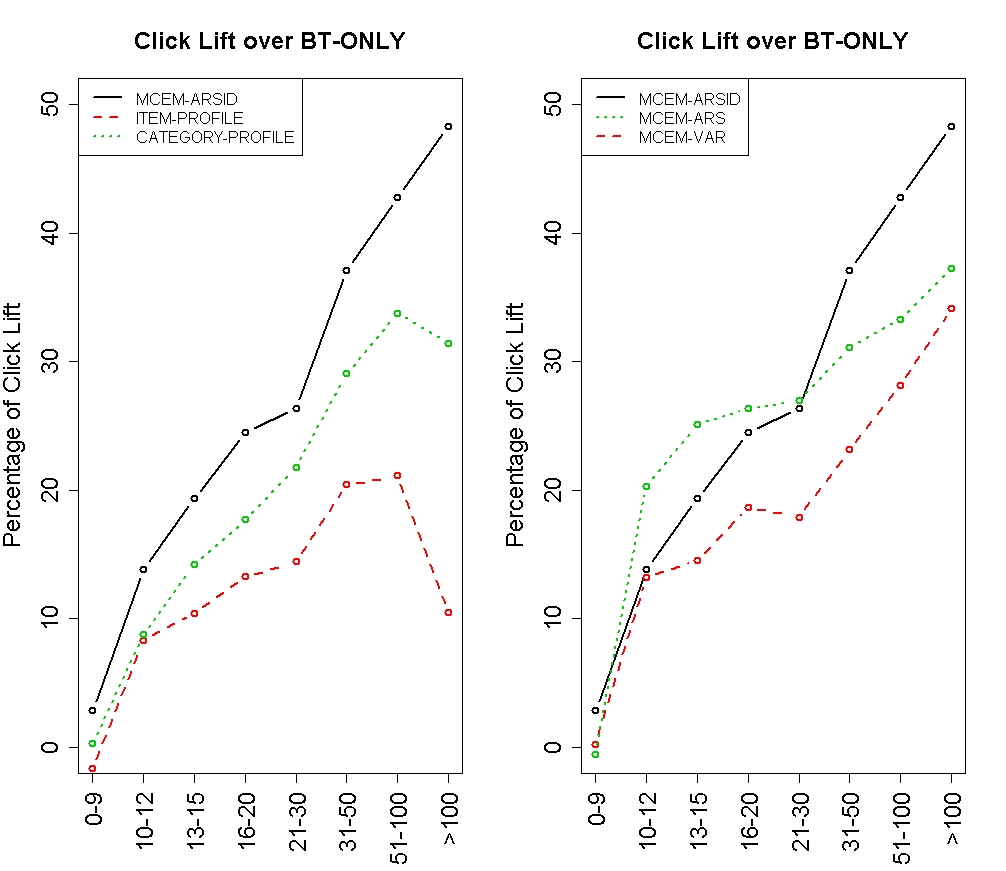}}
\end{center}
\caption{The click lift over the user behavior covariate (BT) only model for different user segments. The segments are created from the number of clicks in the training data.}
\label{figure::ctrlift}
\end{figure}

To further investigate
the performance of algorithms in different segments of warm-start users based on user activity levels on Today module in training period,
we look at click-lifts by Today module activity levels in Figure~\ref{figure::ctrlift}. We split the users in the test data into several segments by their number of clicks in the training data. As expected, we see a near-monotone trend; users with more activity
are personalized better by using their prior Today Module activity data. From Figure \ref{figure::ctrlift} we observe that MCEM-ARSID is uniformly better than CATEGORY-PROFILE and ITEM-PROFILE over all the user segments. Comparing performance of MCEM-ARSID, MCEM-ARS and MCEM-VAR we find the MCEM-VAR to be quite inferior to MCEM-ARS and MCEM-ARSID.

\begin{figure}[ht]
\begin{center}
\centerline{\includegraphics[width=7.5cm]{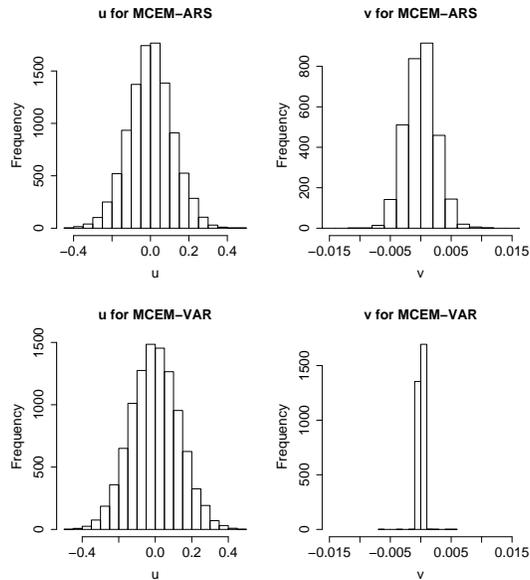}}
\end{center}
\caption{The histogram of the fitted $\bm{u}_i$ and $\bm{v}_j$ after 30 iterations of the MCEM step for \VAR and \ARS, both with 10 factors and 400 partitions.}
\label{figure::var-shrink}
\end{figure}
\parahead{Potential issue with variational approximation}
To investigate issues with MCEM-VAR with data sparsity, we examine the random effect estimates in Figure~\ref{figure::var-shrink}
which shows the histograms of the fitted $\bm{u}_i$ and $\bm{v}_j$ after 30
EM iterations for \VAR and \ARS, both with 10 factors
and 100 partitions.  While the fitted user random effects for both \VAR and
\ARS are in the similar scale, the item random effects produced by the variational
approximation is approximately one order of magnitude smaller than
those produced by \ARS. This phenomenon is in fact surprising and shows that
\VAR tends to over-shrink the random effect estimates when fitting rare response. Similar phenomenon has been observed by \cite{zhang2008fast}. This
explains why the performance of \VAR deteriorates as the binary
response gets rare. It seems that the variational approximation leads to too much
shrinkage when working with rare response.

\subsection{Discussion of Results}
The experiments clearly show that regression-based BIRE models for binary response and using a divide and conquer stategy to scale
the method in a Hadoop framework involves several subtle issues. For scenarios where we can fit the model using
a single machine, all methods work equally well on balanced binary response, the case widely studied in prior work. 

For highly imbalanced data, \VAR tends
to deteriorate, we do not recommend its use in such scenarios. SGD works well provided that the learning rates and regularization parameters are
tuned carefully, hence we do not recommend its use unless such tuning is undertaken seriously. Even after tuning, it is inferior to MCEM methods so
we recommend using MCEM if possible. For single machine MCEM, imposing positivity constraints in \ARSID hurts performance slightly since it adds additional constraints.
Therefore, we do not recommend it, instead we recommend fitting \ARS. 

The story is totally different when fitting map-reduce with divide and conquer. Since the BIRE
models are multi-modal, each partition may converge to a very different regression estimate so that simple average of the regression coefficients leads to poor performance. Here, we highly recommend
making all the efforts to impose identifiability through \ARSID and synchronizing the initializations. We also recommend using the ensemble trick since
it only uses the E-step and does not add too much to the computations. We discourage the use of \VAR since it breaks quite spectacularly with high sparsity.

\section{Conclusion}

In this paper, we introduced the adaptive rejection sampling (ARS) to
our probabilistic regression-based bilinear random effects (BIRE) modeling framework to handle data sets
with binary response in a better way.  We note that data with binary
response is common in web applications such as content optimization
and computational advertising.  We also extended our BIRE model fitting methods to handle large data sets using
Map-Reduce. By extensive experiments on benchmark datasets and the
Yahoo! FrontPage Today Module data sets, we show that our model and
fitting algorithms are stable and can significantly outperform
variational approximation proposed by
\cite{agarwal2009regression,gmf} and several other baselines. We also notice that carefully handling idenfiability issues have crucial impact on the BIRE model performance while handling large-scale data sets using Map-Reduce.

\section{Acknowledgment}

We thank Markus Weimer from Yahoo! Labs for providing the code of fitting matrix factorization using stochastic gradient descent and continuously giving us technical support for understanding and running the code. We are also very grateful to Andrew Cron from Duke University to help us embed the centering idea into our parallel MCEM-ARS fitting algorithm.

\small{
\bibliographystyle{natbib}
\bibliography{paper}
}

\end{document}